\documentclass[utf8]{frontiersSCNS} % for Science, Engineering and Humanities and Social Sciences articles
\usepackage{url,hyperref,lineno,microtype,subcaption}
\usepackage[onehalfspacing]{setspace}

\usepackage{booktabs,multirow}
\usepackage{tabularx}

\def\keyFont{\fontsize{8}{11}\helveticabold }
\def\firstAuthorLast{Tetteh {et~al.}} %use et al only if is more than 1 author
\def\Authors{Giles Tetteh\,$^{1,*}$, Fernando Navarro\,$^{1}$, Johannes Paetzold\,$^{1}$, Jan Kirschke\,$^{2}$, Claus Zimmer\,$^{2}$, and Bj\"orn H. Menze\,$^{1,3}$}
\def\Address{$^{1}$Department of Computer Science, TU M\"unchen, TU M\"unchen, Germany.\\
$^{2}$Neuroradiology, Klinikum Rechts der Isar, TU M\"unchen, TU M\"unchen, Germany.\\
$^{3}$Department of Quantitative Biomedicine, University of Zurich, Switzerland.}
\def\corrAuthor{Giles Tetteh}

\def\corrEmail{giles.tetteh@tum.de}

\begin{document}
\onecolumn
\firstpage{1}

\title[Collateral Flow Grading]{A Deep Learning Approach to Predicting Collateral Flow in Stroke Patients Using Radiomic Features from Perfusion Images} 

\author[\firstAuthorLast ]{\Authors} %This field will be automatically populated
\address{} %This field will be automatically populated
\correspondance{} %This field will be automatically populated

\extraAuth{}% If there are more than 1 corresponding author, comment this line and uncomment the next one.
%\extraAuth{corresponding Author2 \\ Laboratory X2, Institute X2, Department X2, Organization X2, Street X2, City X2 , State XX2 (only USA, Canada and Australia), Zip Code2, X2 Country X2, email2@uni2.edu}

\maketitle

\begin{abstract}
Collateral circulation results from specialized anastomotic channels which are capable of providing oxygenated blood to regions with compromised blood flow caused by ischemic injuries. The quality of collateral circulation has been established as a key factor in determining the likelihood of a favorable clinical outcome and goes a long way to determine the choice of stroke care model - that is the decision to transport or treat eligible patients immediately.

Though there exist several imaging methods and grading criteria for quantifying collateral blood flow, the actual grading is mostly done through manual inspection of the acquired images. This approach is associated with a number of challenges. First, it is time consuming - the clinician needs to scan through several slices of images to ascertain the region of interest before deciding on what severity grade to assign to a patient. Second, there is a high tendency for bias and inconsistency in the final grade assigned to a patient depending on the experience level of the clinician.

We present a deep learning approach to predicting collateral flow grading in stroke patients based on radiomic features extracted from MR perfusion data. First, we formulate a region of interest detection task as a reinforcement learning problem and train a deep learning network to automatically detect the occluded region within the 3D MR perfusion volumes. Second, we extract radiomic features from the obtained region of interest through local image descriptors and deniosing auto-encoders. Finally, we apply a convolutional neural network and other machine learning classifiers to the extracted radiomic features to automatically predict the collateral flow grading of the given patient volume as one of three severity classes - no flow (0), moderate flow (1), and good flow (2).

Results from our experiments show an overall accuracy of 72\% in the three class prediction task. With an inter-observer agreement of 16\% and intra-observer agreement of 74\% in a similar experiment, our automated deep learning approach demonstrates a performance comparable to expert grading, is faster than visual inspection and eliminate the problem of grading bias.

\tiny
 \keyFont{ \section{Keywords:} collateral flow, radiomics, perfusion, reinforcement learning, image descriptors, angiography, auto-encoder, deep learning}
\end{abstract}

\section{Introduction}
Collateral circulation results from specialized anastomotic channels which are present in most tissues and capable of providing nutrient perfusion to regions with compromised blood flow due to ischemic injuries caused by ischemic stroke, coronary atherosclerosis, peripheral artery disease, and similar conditions and diseases \citep{Faber2014}. Collateral circulation helps to sustain blood flow in the ischaemic areas in acute, subacute or chronic phases after an ischaemic stroke or transient ischaemic attack \citep{Liu2018}. The quality of collateral circulation has been convincingly established as a key factor in determining the likelihood of successful reperfusion and favorable clinical outcome \citep{Ginsberg2017}. It is also seen as one of the major determinants of infarct growth in the early time windows which is likely to have an impact on the chosen stroke care model – that is the decision to transport or treat eligible patients immediately.

A high number of imaging methods exist to assess the structure of the cerebral collateral circulation and several grading criteria have been proposed to quantify the character of collateral blood flow. However, this grading
is mostly done through visual inspection of the acquired images which introduces two main challenges.\\
First, there are \textit{bias and inconsistencies in the current grading approaches}: There is a high tendency of introducing bias in the final grade assigned to a patient depending on the experience level of the clinician. Also, there is inconsistency in the grade assigned by a particular clinician at different times for the same patient. This inconsistencies are quantified at 16\% interobserver agreement and 72\% intraobserver agreement respectively in a similar study by \cite{BenHassen2018}.\\
Second, \textit{grading is time consuming and tedious}: Aside the problem of bias prediction, it also takes the clinician several minutes to go through the patient images to first select the correct image sequence, detect the region of collateral flow and then to be able to assign a grading – a period of time which otherwise could have been invested into the treatment of the patient.

In this work, we analyze several machine learning and deep learning strategies that aim towards automating the process of collateral circulation grading. We present a set of solutions focusing on three main aspects of the task at hand.\\
First, \textit{the region of interest needs to be identified}. We automate the extraction of the region of interest from the patient images using deep reinforcement learning. This is necessary in achieving a fully automated system which will require no human interaction and also in saving the clinician the time spent on performing this task.\\
Finally, \textit{region of interest needs to be processed and classified}. We consider various feature extraction schemes and classifiers suitable for the task described above. This helps to extract useful image features, both learned and hand crafted, which are relevant for the classification task. We test digitally subtracted angiography (DSA) based collateral flow grading from MR perfusion images in this task. This saves time required in choosing the right DSA sequence from the multiple DSA sequences acquired and helps achieve a fully automated system.

\subsection{Prior work and open challenges}
\subsubsection{Imaging criteria for cerebral collateral circulation}
Imaging methods for assessing cerebral collateral flow can be grouped under two main classification schemes which are invasive vs non-invasive and structural vs functional imaging. Structural imaging methods provides information about the underlying structure of the cerebral collateral circulation network. Some of the commonly used structural imaging modalities are traditional single-phase computed tomography angiography (CTA), time-of-flight magnetic resonance angiography (TOF-MRA), and digitally subtracted angiography (DSA) among others. Other imaging modalities has been used in clinical practice and relevant research areas in accessing the structure of the cerebral collateral circulation and discussed in \citet{Liu2018,Mcverry2011,Martinon2014}. DSA is the gold standard for assessing the collateral flow, however due to the associated high cost and invasive nature, other non-invasive methods like CTA and MRA are commonly used \citep{Liu2018}.

Functional imaging methods are used to assess the function of the underlying cerebral collateral circulation. Single-photon emission CT (SPECT), MR perfusion and positron emission tomography (PET) are examples of imaging methods which provides functional information about the cerebral collateral flow. MR perfusion imaging is often complimented by a post-processing step to extract parametric information. Very common parametric information include the \textit{time to peek} ($\text{T}_{max}$) – time taken for the blood flow to reach it’s peek (max) at a given region in the brain, \textit{relative blood flow} (rBF) – volume of blood flowing through a given brain tissue per unit of time, and \textit{relative blood volume} (rBV) – volume of blood in a given brain tissue relative to an internal control (e.g. normal white matter or an arterial input function). Functional imaging are sometimes combined with structural imaging either in a single scanning procedure or separate procedures and they serve as complementing information towards the decision making process. Here, structural imaging is oftentimes used to map the anatomy and probe tissue microstructure.

Together, MRI perfusion and diffusion have evolved as key biomarkers in determining collateralization of stroke patients, and a patient stratificaiton based on these markers has been proposed repeatedly. At the same time, a qualitative DSA based grading is the most reliable approach for evaluating collateralization. 

In this study, we employ parametric information ($T_{max}$ , rBF, rBV) from MR perfusion images of patients with acute ischaemic stroke and predict the three-level DSA based grading of these patient based on this functional information. We hypothesize that the rich information on blood flow visible from MRI perfusion can be used to predict collateral flow in a similar manner to DSA. Moreover, we argue that this approach, using 3D information, may even offer a more reliable biomarker than the interpretation of DSA images. As collateralization patterns are often instable and may underly significant changes in the course of minutes, a second estimate of the activation of collateral flow using MRI - in addition to the subsequent DSA - will always offer better diagnostic information.

\subsubsection{Cerebral collateral flow grading}
The cerebral collateral circulation plays an important role in stabilizing cerebral blood flow when normal blood circulation system is compromised in cases of acute, subacute or chronic ischaemic stroke. The quality of the cerebral collateral circulation system is one of the factors that determines the speed of infarct growth and the outcome of stroke treatment and reperfusion therapies. Poor collateral flow is associated with worse outcome and faster growth of infarcts while good collateral flow is associated with good outcome and slower growth of infarcts in acute stroke treatment \citep{Jung2017}. Due to the important role played by cerebral collateral blood flow, various grading scales and their association with risk factors and treatment outcomes has been discussed extensively in literature.

Several grading system has been proposed for assessing the quality of the collateral circulation network, Among these grading system, the DSA based system proposed by the American Society of Interventional and Therapeutic Neuroradiology/Society of Interventional Radiology (ASITN/SIR) is the most widely accepted scheme. This grading system describes the collateral flow as one of five levels of flow which are; absence of collaterals (0), slow collaterals (1), rapid collaterals (2), partial collaterals (3), and complete collaterals (4) to the periphery of the ischaemic site \citep{Liu2018,Maija2015}. In most studies which uses the ASITN/SIR scheme, the grading scale is merged into three levels -- grades $0-1$ (poor), 2 (moderate) and $3-4$ (good collateral) flow. CTA based systems also have several grading schemes ranging from two (good, bad) to five (absent, diminished $>50\%$, $<50\%$, equal, more) \citep{Maija2015}.

The relationship between pretreatment collateral grade and vascular recanalization has been assessed for consecutive patients who received endovascular therapy for acute cerebral ischemia from two distinct study populations by \citet{Bang2011}. The study showed that 14.1\%, 25.2\% and 41.5\% of patients with poor, good, and excellent pretreatment collaterals respectively achieved complete revascularization. Another study by \citet{Bang2008} on the relationship between MRI diffusion and perfusion lesion indices, angiographic collateral grade and infarct growth showed that greatest infarct growth occurred in patients with both non-recanalization and poor collaterals. \citet{Mansour2013} assessed collateral pathways in acute ischemic stroke using a new grading scale (Mansour Scale) and correlated the findings with different risk factors, clinical outcome and recanalization rate with endovascular management. More research \citep{Mass2009,Tan2009, Mansour2013, Bang2008, Bang2011} has been conducted into the relationship between the cerebral collateral circulation, it’s grading and the clinical outcome of the choice of treatment of acute ischemic stroke and they all confirm a positive association between collateral flow and the success of the outcome.

Due to the crucial role played by collateral circulation, it is a common practice in most clinical procedures to determine the quality of a patient’s collaterals as a first hand information towards the choice of the treatment or care model. This grading is done manually by inspecting patient scans which is time consuming and also introduces some level of bias in the final grade assigned to a patient. \citet{BenHassen2018} evaluated the inter-and intraobserver agreement in angiographic leptomeningeal collateral flow assessment on the ASITN/SIR scale and found an overall interobserver agreement $\kappa = 0.16 \pm 6.5 \times 10^{-3}$ among 19 observers with grades 0 and 1 being associated with the best results of $\kappa = 0.52 \pm 0.001$ and $\kappa = 0.43 \pm 0.004$ respectively. By merging the scales into two classes, poor collaterals (grade 0, 1, or 2), versus good collaterals (grade 3 or 4), the interobserver agreement increased to $\kappa = 0.27 \pm 0.014$. The same study recorded a maximum intraobserver agreements of $\kappa = 0.74 \pm 0.1$ and $\kappa = 0.79 \pm 0.11$ for the ASITN/SIR and dichotomized scales respectively. \citet{McHugh} presented a study on interrater reliability and the kappa statistic as a measure of agreement and recommended a moderate interobserver agreement of $0.60 \leq \kappa \leq 0.79$ as a minimum requirement for medical data and study. These results are evidence of the need to automate the collateral grading process to achieve speed and consistency in the assigned grading.

Methods for automating the grading of collateral flow have not yet been properly explored in literature. \citet{Kersten-Oertel2020} presented an automated technique to compute a collateral circulation score based on differences seen in mean intensities between left and right cerebral hemispheres in 4D angiography images and found a good correlation between the computed score and radiologist score ($r^2 = 0.71$) and good separation between good and intermediate/poor groups. \citet{Grunwald2019} used a machine learning approach to categorise the degree of collateral flow in 98 patients who were eligible for mechanical thrombectomy and generated an e-CTA collateral score (CTA-CS) for each patient. The experiments showed that the e-CTA generated improved the intraclass correlation coefficient between three experienced neuroradiologists from 0.58 (0.46-0.67) to 0.77 (0.66-0.85, p = 0.003). In this study, we explore machine learning and deep learning methods in collateral flow grading using parametric information extracted from MR perfusion data of acute ischemic stroke patients.

\subsubsection{Reinforcement learning for medical imaging}
Defining the region of interest is often the first step in most image-based radiomics pipeline. This is the case because the full patient scans normally include artifacts and other information which are irrelevant and can affect the final outcome of the study. Most pipelines therefore propose a manual localization of ROI as a preprocessing step. However, it is crucial to define the region of interest in an automated and reproducible fashion in other to achieve a fully automated pipeline and a real clinical translation. We propose a reinforcement learning approach for localization of the region of interest due to speed and lower requirements for training data as compared to other supervised learning approaches.

Reinforcement learning (RL) has become one of the most active research areas in machine learning and involves the training of a machine learning agent to make a sequence of reward based decisions towards
the achievement of a goal through interaction with the environment. The idea of RL has been long applied in the field of robotics for robot vision and navigation \citep{Pauli2001,Desouza2002,Bonin-Font2008} before the topic became very popular among the image processing society. RL has been used in the general field of computer vision mainly for object detection \citep{Peng1996,Peng1998,Taylor2004}, image segmentation \citep{Sahba2006,Sahba2007}, and image enhancement \citep{Shokri2003,Sahba2005,Tizhoosh2006}. However, in medical imaging RL is still on the research phase with very high posibilities. \citet{Netto2008} presented an overview of medical imaging applications applying reinforcement learning with a detailed illustration of a use case involving lung nodules classification which showed promising results. \citet{Sahba2006} implemented a RL based thresholding for segmenting prostate in ultrasound images with results which showed high potential for applying RL in medical image segmentation.\\
In this work we apply deep reinforcement learning, a variant of RL which combines the power of deep learning and reinforcement learning, to detect a rigid-sized cube around the occluded region in acute ischemic stroke patient scan towards the prediction of cerebral collateral flow grading. This step is necessary to automate the detection of the occluded region which improves the accuracy of the prediction. Saving the time spent on this task and ensuring that the proposed methodology is fully automated.

\section{Methodology}
\begin{figure}[!th]
\includegraphics[width=\textwidth]{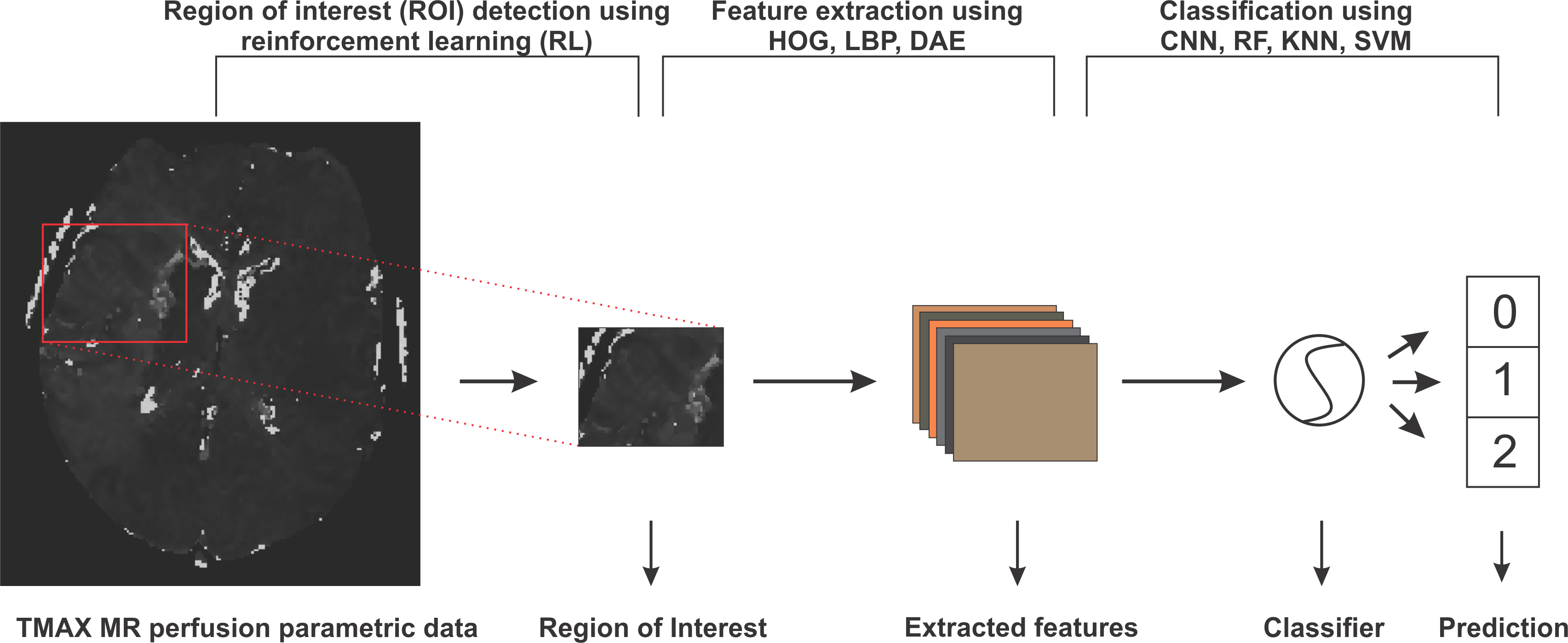}
\caption{An overview of the steps involved in predicting collateral flow grading from MR perfusion parametric data. First step involves region of interest detection using reinforcement learning, followed by
histogram of gradient (HOG), local binary pattern (LBP), denoising autoencoder (DEA) feature extraction schemes and then classification step which uses random forest (RF), K-nearest neighbor (KNN), support
vector machine (SVM), and convolutional neural network (CNN) classifiers}\label{fig:overview}
\end{figure}
In this section we will discuss the details of the steps we employed in predicting the collateral flow grading from MR perfusion data. Figure \ref{fig:overview} shows an overview of the main steps involved in the classification process. The first step is the detection of the region of interest (ROI) using reinforcement learning. This step helps to narrow down the classification task to only the area which has been occluded from normal blood flow. The second step deals with extracting features from the ROI. Finally, we feed the extracted features to a set of classifiers to obtain the collateral grading for the given patient data. Details of these steps and discussed in the next subsections.

\subsection{Deep reinforcement learning for region of interest (ROI) detection}\label{sec:rl}
The idea of reinforcement learning includes an artificial agent which is trained to interact with an environment through a sequence of reward based decisions towards a specific goal. At every time step $t$, the agent takes into account its current state $s$ and performs and action $a$ in a set of actions $A$ and receives a reward $r$ which is a measure of how good or bad the action $a$ is towards the achievement of the set goal. The aim of the agent, which is to find an optimal policy (set of states, actions and rewards) that maximizes the future reward, can be formulated as a Markov Decision Process. Since Markov Decision Process involves a large amount of possible decision points which are normally not fully observable, RL approximates the optimal decision function by iteratively sampling from the set of policies through a process known as Q-learning.

\subsubsection{Q-Learning}
At time point $t$ and state $s$, let $\pi = {a_i}^{t+T}_{i=t}$ be a policy which is a sequence of actions needed by the agent to move from the current state $s$ to the target. Lets $Q_t$ be a future discounted reward function such that
\begin{equation}
Q_t(s,\pi) = \sum_{i=t}^{t+T} \gamma^{i-t}r_{\pi i}\label{eqn:reward-fnc},
\end{equation}
where $r_{\pi i}$ is the reward associated with the action $a_i$ of policy $\pi$ at time $t = i, \gamma \in [0, 1]$ is the future reward discounting factor and $T$ is the number of steps needed to reach the target by the chosen policy $\pi$. At any time step $t$ the optimal policy $\pi^*$ is the policy which maximizes the expected value of $Q_t$. This can be represented by an action-value function $Q^{\ast}_t(s)$ define by
\begin{equation}
\pi^* = Q^*_t(s) = \max_{\pi}\mathbb{E}[Q_t(s,\pi)]
\end{equation}
The optimal value function $Q^{\ast}_t(s)$ obeys the Bellman equation, stating that if the optimal value $Q^{\ast}_{t+1}(s)$ of the next state is known for all possible policies $\pi$, then the optimal behaviour is to select the policy $\pi^{\ast}$ which maximizes the expected value of $r_{\pi t} + Q_{t+1}(s, \pi)$ (which follows from setting $i=t$ in \eqref{eqn:reward-fnc}). The action-value function can therefore be estimated recursively as
\begin{equation}
Q^*_t(s) = \max_{\pi}\mathbb{E}[r_{\pi t} + Q_{t+1}(s,\pi)] \label{eqn:action-value-fnc}
\end{equation}
If the problem space is small enough then the set of policies and state can be fully observed and \eqref{eqn:action-value-fnc} can be used to determine the optimal policy towards the target. However, in most cases the problem space is too complex to explore and hence evaluating the future-reward for all possible policies is not feasible.
$Q^{\ast}_t(s)$ is therefore approximated by a non-linear deep network $Q^{\ast}(s, \theta)$ with a set of parameters $\theta$ resulting in what is know as deep Q-learning \citep{Mnih2015}.

\subsubsection{Agent state, action definition, and reward function}
Given a 3-D scan as the agent’s environment, a state $s$ is represented by $(s_x, s_y, s_z)$ which is the top-left corner of a $(64 \times 64 \times 64)$ cube contained in the 3-D scan. We adopt an agent history approach which involves feeding the last four states visited by the agent to the network to prevent the agent from getting stuck in a loop. Since we have a fixed sized cube as a state our agent’s set of six actions $\{m_u, m_d, m_l, m_r, m_f, m_b\}$ is made up of only movements up, down, left, right, forward, and backward respectively which enables the agent to visit all possible locations within the volume. Agent’s reward for taking an action $a$ is a function of the intersection over union (IoU) of the target state $s^\ast$ and the state before $(s_{ab})$, and after $(s_{aa})$ taking the action. This is given by
\begin{equation}
R_a(s_{aa},s_{ab}) = sign[IoU(s_{aa}, s^{\ast}) - IoU(s_{ab}, s^{\ast})]
\end{equation}
where $sign$ is the sign function which returns $-1$ for all values less than $0$ and $1$ otherwise. This leads to a binary reward $(r \in \{-1, 1\})$ scheme which represents good and bad decisions respectively. At training the agent search sequence is terminated when the IoU of the current agent’s state and the target state is greater than or equal to a predefined threshold $\tau$ . At test time the agent is terminated when a sequence of decisions leads to an oscillation (as proposed by \citet{Alansary2019}), that is when the agent visit one state back and forth for a period of time.

\subsection{Feature extraction and classification}\label{sec:feature-extraction}
Feature extraction methods are used in many machine learning tasks to either reduce the dimension of the problem or to extraction information from the raw input which would otherwise not be easily extracted by the underlying classifier. In this work, we extract two main classes of features which are learned features through denoising auto-encoder (DAE), and local image descriptors made up of histogram of oriented gradients (HOG) and local binary pattern (LBP).
\subsubsection{Denoising auto-encoder}\label{sec:dae}
An auto-encoder is an unsupervised deep learning method used for dimension reduction, feature extraction, image reconstruction or deniosing and sometimes also used as a pre-training strategy in supervised learning networks. An auto-encoder is made up of two parts an encoder $\Phi:\mathcal{X} \rightarrow \mathcal{F}$ which maps an image $x \in X$ to $f_x \in \mathcal{F}$ in the features domain and a decoder $\Psi : \mathcal{F} \rightarrow \mathcal{X}$ which maps a feature set $f \in \mathcal{F}$ to $x_f \in X$. The full auto-encoder is therefore a composite function of the form $\Psi \circ \Phi : \mathcal{X} \rightarrow \mathcal{X}$. Let $\widehat{y} = \Psi(\Phi(x))$ for a given input image $x \in \mathcal{X}$ then the learning process of auto-encoder involves finding a pair of $\{\Phi, \Psi\}$ such that such that $\widehat{y}_i = x_i$ for all $x_i \in \mathcal{X}$. The encoder $\Phi$ then becomes the feature extractor which is used for extracting the needed features.

\begin{figure}[!th]
\includegraphics[width=\textwidth]{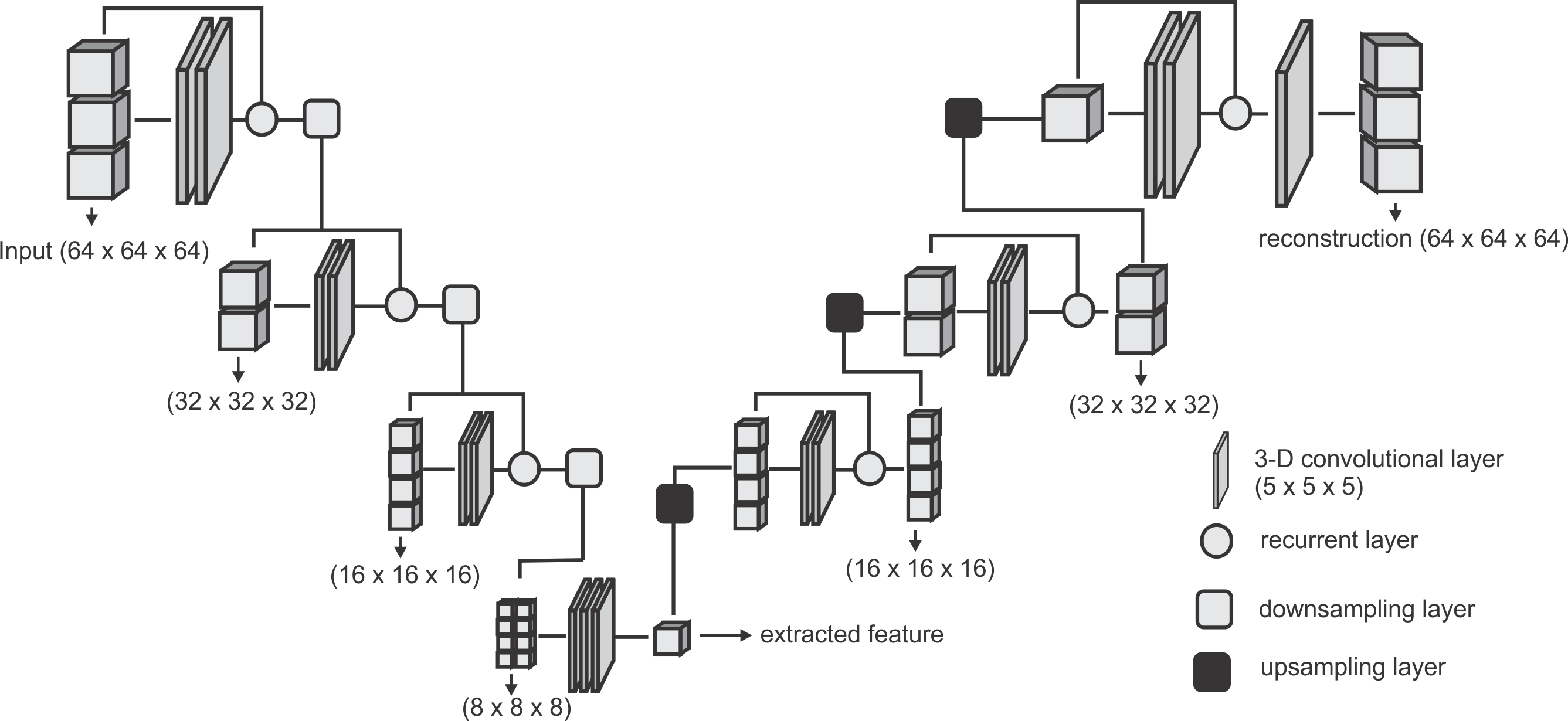}
\caption{Network architecture used for extracting the DAE features. Downsampling layer is a convolution with a stride of $(2 \times 2 \times 2)$ which downsamples the input volume to half of the size on every axes. Upsampling layer is a transposed convolution with a stride of $(2 \times 2 \times 2)$ which doubles the size of the input on every axes.}\label{fig:dae}
\end{figure}

If the function $\Phi$ is invertible, then the learning process can lead to a trivial solution by just choosing $\Psi$ to be the inverse of $\Phi$ and $\Psi \circ \Phi$ becomes an identity function leading what is known as identity-function risk. To prevent this, the input image $x$ is first corrupted by adding noise before feeding it to $\Phi$ leading to a denoising auto-encoder. We there have
\begin{equation}
\widehat{y} = \Psi(\Phi(\tilde{x})), \ \ \ \tilde{x} = \Upsilon(x)
\end{equation}
where $\Upsilon$ is the random image corruption function. We approximate the encoder and decoder by deep CNNs $E(x, \theta_e)$ and $D(f,\theta_d)$ parameterized by $\theta_e$ and $\theta_d$; respectively. Training is done through back-propagating the MSE of the original image $x$ and the reconstructed image $\widehat{y}$ given by
\begin{equation}
\mathcal{L} = \frac{1}{N}\sum_{i=1}^{N}(\widehat{y}_i - x_i)^2
\end{equation}
where $N$ is the number of images in the training set or training batch. Figure \ref{fig:dae} shows an overview of the architecture used for extracting the DAE features.

\subsubsection{Local image descriptors}
We consider two types of local image descriptors - histograms of oriented gradients (HOG) and local binary pattern (LBP). Given a volume $X$, we extract the LBP encoding of each voxel by thresholding
its $3 \times 3 \times 3$ neighbourhood by the intensity value $p^\ast$ of the center voxel which results in $26$ long bits ${b_0, b_1, b_2, ..., b_{25} }$ where $b_i = \{1, \text{if} \ p_i \geq p^\ast, 0 \ \text{otherwise}\}$ and $p_i$ is the intensity value of the $i$th neighbour. We then concatenate the binary encoding to a single binary number $b_0 b_1 b_2 ...b_{25}$ and then into a decimal
value which results in $2^{25}$ possible binary codes. Details of the implementation till this point can be found in \citet{Heikklae2006}. We group the codes into two main classes - uniform codes which have at most two binary transitions and non-uniform codes which have more than two binary transitions. A binary transition is a switch from 0 to 1 or vice versa. For example the codes $0000$, $000111$, $011100$, and $110110$ have zero, one, two, and three transitions respectively. To handle noisy data and to reduce the feature space, we group all the non-uniform codes into one class and add it to the uniform codes resulting in $927$ codes instead of $2^{25}$ . Finally, the histogram distribution of the individual codes are extracted as the LBP feature representation for the volume $X$.

We also explore the HOG feature extractor based on the method proposed in \citet{Klaeser2008}. Given a volume $X$, we quantize gradient orientations over an icosahedron and merge opposite directions in one
bin resulting in 10 gradient orientations. Gradient for each voxel $x_i \in X$ is obtained by convolving the $5 \times 5 \times 5$ neigbourhood of the voxel by gradient filters $k_x$ , $k_y$ , and $k_z$ of the same size, giving us a gradient vector $\overrightarrow{x}_i \in \mathbb{R}^3$. The gradient filters are zero everywhere except for the center columns along the respective axes $k_x (i, 3, 3) = k_y (3, i, 3) = k_z (3, 3, i) = [1, 0, -2, 0, 1]$ for $i \in \{1, 2, ..., 5\}$. The gradient vectors $\overrightarrow{x}_i$ are then projected to the gradient orientations and a histogram representation of these orientations are obtained and used as the HOG feature representation of the volume $X$.

\subsubsection{Classifiers}
\begin{figure}[!th]
\includegraphics[width=\textwidth]{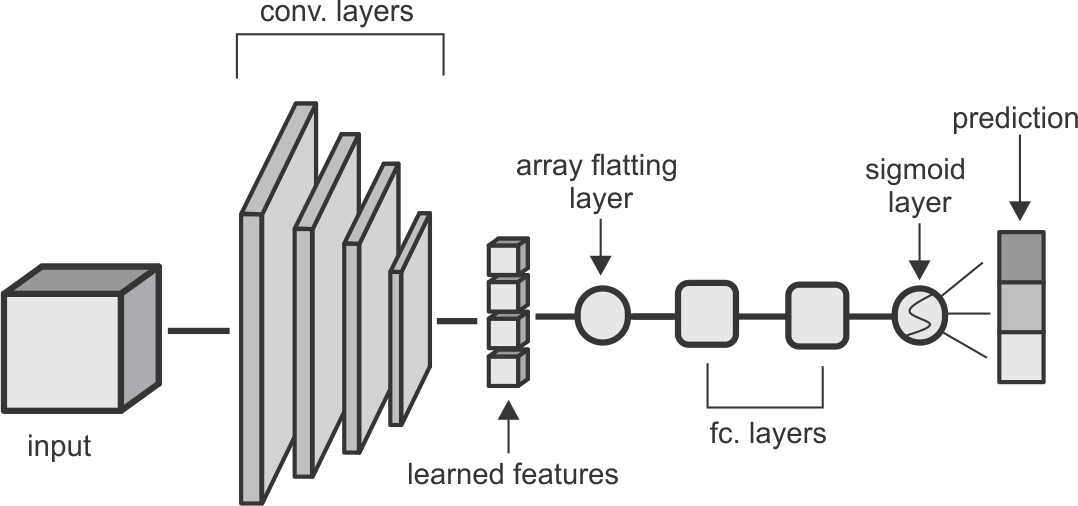}
\caption{CNN architecture used in the classification task. Convolutional layers are made up of $(5 \times 5 \times 5)$ kernels with a stride of $(2 \times 2 \times 2)$ which reduces the volume by half of the input size on each layer. The first two layers extract 2 feature cubes and the last two layers extract 4 feature cubes each. The fully connected layers have 64 and 32 hidden nodes respectively and of the convolutional and fully connected layers is followed by a non-linear hyperbolic tangent (tanh) activation function.}\label{fig:cnn}
\end{figure}
We explore four classifiers on each of the features extracted. We implement Convolutional Neural Network (CNN), Random Forest (RF), Support Vector Machine (SVM) and K-Nearest Neighbour (KNN) classifiers. Our CNN classifier in Figure \ref{fig:cnn} has four convolutional layers, aimed at extracting local image features, followed by two fully connected layers and a sigmoid layer for classification. Each layer is followed by a non-linear hyperbolic tangent (tanh) activation function. For classification based on the HOG, LBP and DEA features, we remove the convolutional layers and feed the features directly to the fully connected layers and then the sigmoid layer for the classification. For the RF, SVM, and KNN classifiers we use the implementation of these classifiers from the Scikit-Learn library in python.

\section{Experiments and results}
\subsection{Patient population and image data}
We test our proposed methods on parametric volumes extracted from MR perfusion data from 183 patients with acute ischemic stroke. Our dataset is made up of three parametric information -- $T_{max}$ volumes which refers to the time taken for the blood flow to reach it’s peek, relative blood flow (rBF) volumes which refers to the volume of blood passing through a given brain tissue per unit of time, and relative blood volume (rBV) defined as the volume of blood in a given brain tissue relative to an internal control (e.g. normal white matter or an arterial input function). Each volume has a resolution of (0.9mm, 0.9mm, 6.5mm) and a dimension of (256, 256, 19) voxels on the sagittal, coronal and axial planes respectively. Ground truth labels are obtained from a trained neuroradiologist, with over ten years experience, who manually investigate the DSA slides of the associated patient and assign one of three labels (0-bad, 1-medium, 2-good) to this patient. We use these labels for a direct 3-class prediction experiment and we also experiment on a risk-stratified nested test where we first predict good - (2) against not good (0, 1) collaterals and then separate the not good class into bad (0) and medium (1) collaterals in a cascaded approach.

\subsection{Preprocessing}
Our image preprocessing involves two main tasks. First, we make our datasets isotropic by applying a B-spline interpolation to the third axis since the first two axes have the same spacing leading to volume with resolution 0.9mm on each plane. This is followed by an extraction of the brain region from the skull using the brain extraction tool (BET) from the ANTS library. The brain extraction is carried out on the $T_{max}$ volumes and the resulting mask is then applied to the rBF and rBV volumes.

\subsection{Region of interest localization}
\begin{figure}[!ht]
\includegraphics[width=0.48\textwidth]{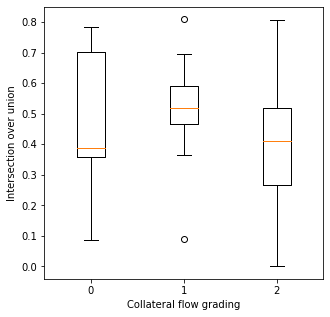}
\includegraphics[width=0.48\textwidth]{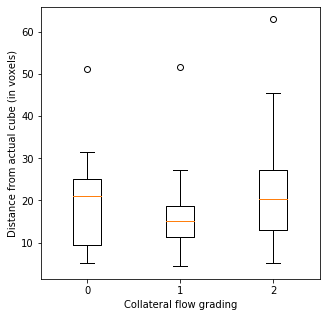}
\caption{Box plots of results from ROI detection task. Left is the the intersection over union (IoU) ratio between the prediction and the ground truth over the three classes. Right is the euclidean distance between the predicted center point and ground truth center point. From the distributions it is clear that its easy to detect the ROI in the poor collateral flow class (class 0) compared to the good collateral flow class (class 2). This can be explained by the fact that in good collateral flow cases $T_{max}$ shows uniform values in the whole
volume.}\label{fig:box-plots}
\end{figure}

After the preprocessing step we extract the occluded regions as the region of interest (ROI) using the reinforcement learning architecture described in Section \ref{sec:rl}. We adopt the network architecture from \citet{Alansary2019} with modifications proposed in \citet{Navarro2020}. A stopping criteria threshold $\tau = 0.85$ is used during training -- that is an intersection over union (IoU) value of greater than or equal to $0.85$ implies that the region of interest is detected. We perform the ROI detection task on the $T_{max}$ volumes since the occluded regions are easier to detect in these volumes. For each volume, we select $20$ starting cubes of size $(64 \times 64 \times 64)$ at random and run the agent till the stopping criteria is reached. We then aggregate the results from the $20$ different runs to get the prediction of the final ROI. After getting the region we extract the mirror of the ROI by reflecting the ROI on the opposite side of the brain and use it as an additional feature. This results in 6 cubes per patient (e.i. two volumes each from $T_{max}$ , rBF and rBV volumes). Qualitative and quantitative results from the region of interest extraction can be found in Table \ref{tab:roi} and Figures \ref{fig:roi} and \ref{fig:box-plots}. From the box plots in Figure \ref{fig:box-plots}, it is evident that the region of interest detection was more successful in the poor collateral flow class than in the good collateral flow class. This can be explained by the fact that in cases of good collateral flow, there is uniform distribution of the $T_{max}$ value within the occluded region and its neighbourhood making it had for the RL agent to detect the ROI. From Figure \ref{fig:roi} we can observe that in most cases the ground truth does not coincide with the whole of the occluded region (e.g. column (b)) and hence though the predicted ROI does not completely overlap the ground truth it still contains other part of the occluded region which is not in the ground truth and it is therefore sufficiently accurate for the classification task.
\begin{table}
\caption{Quantitative results from region of interest detection task. IoU refers to the intersection over
union ratio between the prediction and the ground truth. Center point displacement is the euclidean
distance between the predicted center point and ground truth center point.}\label{tab:roi}
{
\newcolumntype{b}{X}
\newcolumntype{s}{>{\hsize=.70\hsize\raggedleft\arraybackslash}X}
\begin{tabularx}{\textwidth}{bbssss}
  \toprule
  Type & Class & Mean & Std & Max & Min  \\
  \midrule
       &  0	& 0.49 & 0.22 & 0.79 & 0.08 \\
IoU    &  1	& 0.52 & 0.14 & 0.81 & 0.09 \\
       &  2	& 0.42 & 0.21 & 0.81 & 0.04 \\
  \midrule
Center points   &  0  & 20 & 13 & 51 & 5 \\
displacement    &  1  & 17 &  9 & 52 & 4 \\
(in voxels)     &  2  & 23 & 14 & 63 & 5 \\
  \bottomrule
\end{tabularx}
}
\end{table}

\begin{figure}[!h]
\includegraphics[width=\textwidth]{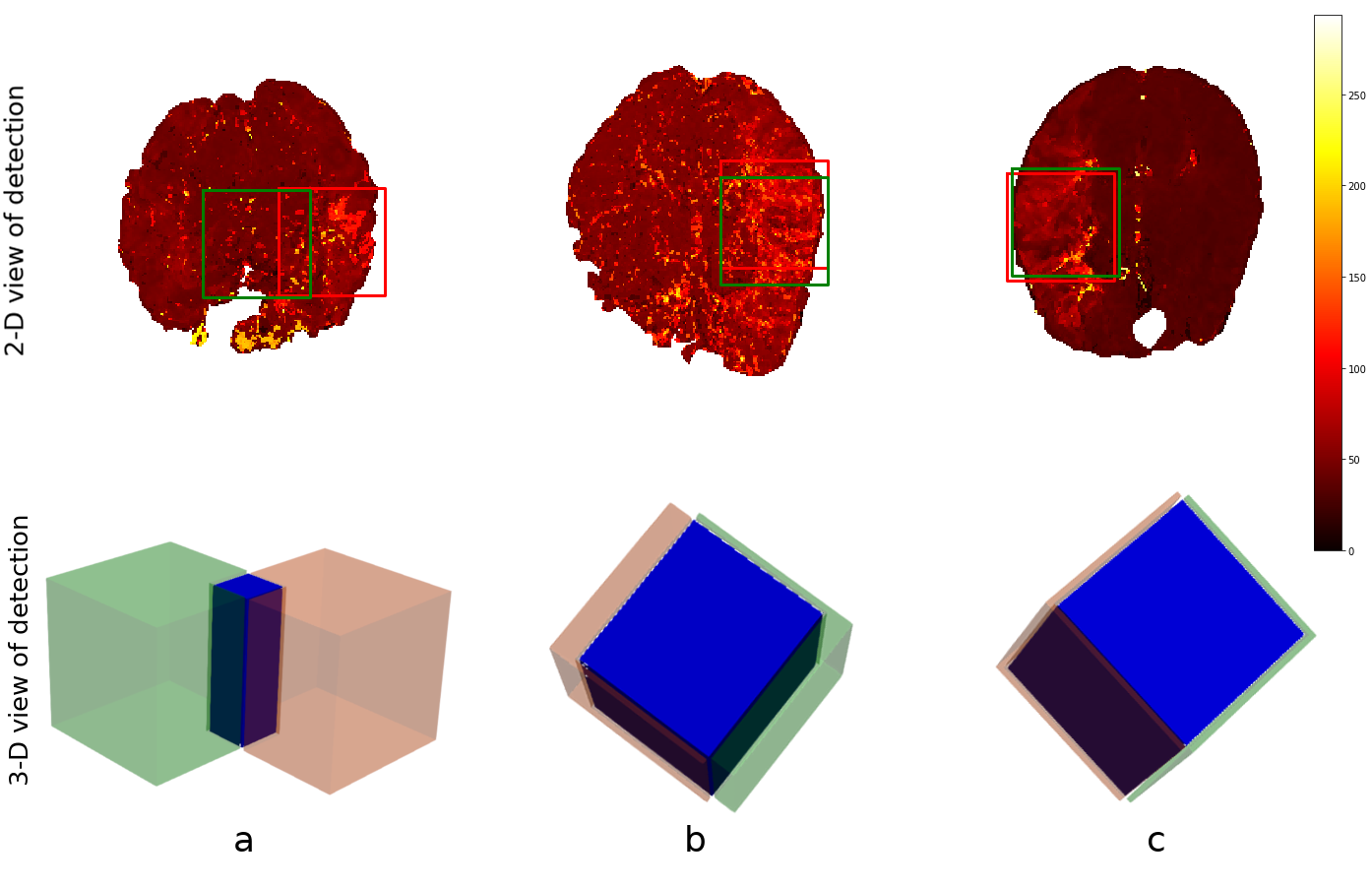}
\caption{Qualitative results from the RIO detection task. Top row is the axial view of the ground truth (in red) and the prediction (in green). Bottom row is a 3-D visualization of the ground truth cube (in red), predicted cube (in green) and the intersection between the two (in blue). Column (a) corresponds to the worse prediction in our test set while column (c) refers to the best in terms of IoU. In column (b), we can observe that though the overlap is not perfect the prediction still contains some part of the occluded region which is not in the ground truth. This implies that though we have poor scores we still have good ROI detection which can be used for the classification task.}\label{fig:roi}
\end{figure}

\subsection{Classification}
\subsubsection{Feature representations}
\begin{figure}[!th]
\includegraphics[width=\textwidth]{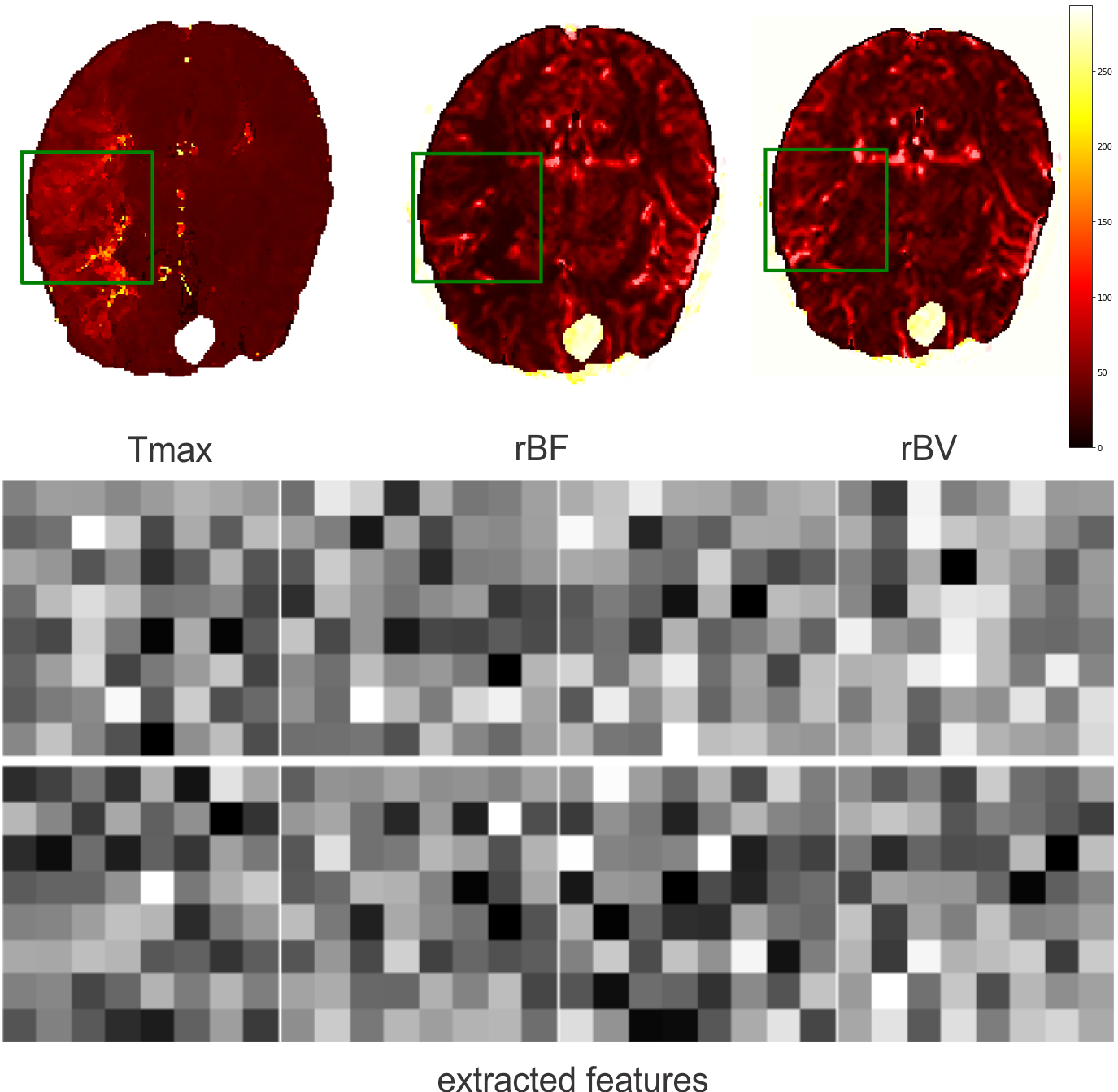}
\caption{Sample of extracted features using denoising auto-encoder. Green boxes refers to the predicted cubes in the patient volumes. Each feature square refers to a slice in the $(8 \times 8 \times 8)$ feature cube.}\label{fig:extracted-features}
\end{figure}

In total three features are extracted together with the actual extracted cube and its mirror cube. We learn features automatically through an unsupervised denoising auto-encoder. The network takes the extracted ROI cubes from the $T_{max}$ , rBV and rBF volumes as three channels and produces a single channel feature set of size $(8 \times 8 \times 8)$. We normalize the cubes individually into the range [0, 1] before feeding them to the network. Figure \ref{fig:extracted-features} shows a sample of the extracted auto-encoder features laid out in a 2-D grid.

\begin{figure}
\includegraphics[width=0.5\textwidth]{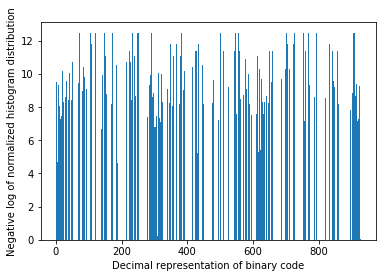}
\includegraphics[width=0.5\textwidth]{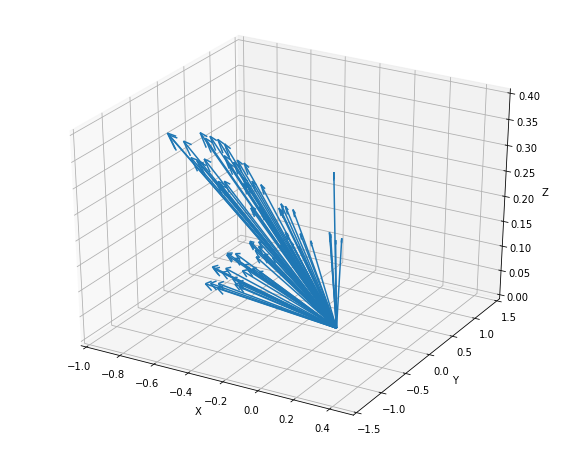}
\caption{Sample of extracted features using Local Binary Pattern (LBP) on the left and Histogram of Oriented Gradients (HOG) on the right. Bar heights from LBP features represent the frequency of a given pattern and the position on the x-axis is the decimal representation of the binary pattern. Arrows direction in HOG features are the gradient vectors and the length of the arrows represent the frequency of the given gradient. The LBP features show a uniform distribution of the extracted patterns with no dominant pattern. The HOG features on the other hand show evidence of high gradients in the extracted region of interest.}\label{fig:lbp-hog-features}
\end{figure}

For HOG features we extract 10 features each for the three parametric volumes and concatenate them into a vector of length 30 for the classification task. Figure \ref{fig:lbp-hog-features} shows a sample of the extracted HOG features for a patient for the three input channels. Finally LBP features are extracted using the method described in Section \ref{sec:feature-extraction}. Here we combine all the three channels and run the histogram over the three channels which results in $927$ feature vector as explained in Section \ref{sec:feature-extraction}. Figure \ref{fig:lbp-hog-features} shows a sample of the extracted LBP features from our dataset.

\subsubsection{Classifier training}
We handle the collateral flow classification through two main approaches -- a three-class multi-label classification task where we predict the label in one step, and a cascaded approach where we predict a binary label of classes (0, 1) against 2 in a first step and a second step where we separate the class 0 from 1.  We implement our CNN architecture using the keras library in python with tensorflow as backend. Random forest, suppor vector machine and K-nearest neighbor classifiers were implemented using the scikit-learn library in python. We setup our experiments as follows:\\
\textit{\textbf{CNN classifier}}: for the CNN classifier we use a weighted categorical crossentropy with a weight of $\frac{1}{|k|}$ for each class $k$ in the training set. A stochastic gradient descent optimizer with a learning rate of $0.001$, decay of $1e^{-6}$ and momentum of $0.9$ is used to fine-tune the network parameters at $20$ epochs.\\
\textit{\textbf{K-Nearest neighbor classifier}}: we conduct preliminary experiment with grid search to know which parameters will work best. For our final experiment we use $k = 3$ neighbors with uniform weights, a leaf-size of $30$ and the minkowski metric.\\
\textit{\textbf{Random forest classifier}}: after the initial grid search experiment we implement the classifier with 200 estimators and the Gini impurity function is used to measure the quality of a split.\\
\textit{\textbf{Support vector machine classifier}}: we use a regularization parameter $C = 10$, a third degree polynomial kernel, a balanced class weight and a tolerance of $1e^{-3}$ for the stopping criterion.

\subsubsection{Classification results}
We experiment various combinations of the features extracted in the previous experiments and classifiers discussed and present the results in Table \ref{tab:classification}. Due to limitation in the size of the dataset, we adopt a 5-fold cross-validation approach instead of a training-validation-test splitting approach and report the average scores over the accuracy in the individual validations.

The results in Table \ref{tab:classification} show that the region of interest extraction step helps improve the results in all classification methods. This can be verified by comparing the RAW column with ROI column in Table \ref{tab:classification}. Also by adding the mirror of the occluded region to the extracted ROI we achieve improved results in most of the classifiers with performance falling in classifiers like KNN and SVM due to the associated increase in the dimension of data introduced by the mirror of the ROI. The cascaded method shows higher accuracy in almost all the classifier-feature combinations when compare to the direct three-class prediction. This can be explained by the distribution of classes in the dataset. That is, for the cascaded approach we have a fairly balanced data when we combine poor and moderate flow against good collateral flow which is not the case with the direct three-class multi-label prediction approach. It therefore suggest that in cases where we have highly imbalanced class distributions a multi-label classification might perform poorly. The overall performance of CNN was better than the other machine learning classifiers and can be explained by the fact that the convolutional layers of the CNN architecture extract features while paying attention to the class of the input data. This makes the feature extraction process more efficient than the other feature extraction schemes which have no knowledge of the underlying label of the input data at the time of extracting the features. Again CNN with only ROI data performs slightly better than with the mirror of the ROI (72\% vs 70\% in Table \ref{tab:classification}) and this can also be explained by the fact that the CNN used in our experiments is fairly shallow and hence could not handle the additional feature dimensions introduced by the mirrored images.

\begin{table}[!th]
\caption{Results from experiments on collateral flow grading. RAW features refers to the three parametric volumes ($T_{max}$ , rBF, and rBV) after skullstriping. ROI refers to the corresponding cubes extracted from the parametric volumes based on the ROI prediction and ROI+M is the ROI combined with its mirror cube in the opposite side of the brain. Other features (DEA, HOG and LBP) are all extracted from the ROI cubes.}\label{tab:classification}
{
\fontsize{10pt}{20pt}
\selectfont
\newcolumntype{b}{X}
\newcolumntype{m}{>{\hsize=.45\hsize\arraybackslash}X}
\newcolumntype{s}{>{\hsize=.55\hsize\raggedright\arraybackslash}X}
\begin{tabularx}{\textwidth}{mmssssss}
  \toprule
  Type & Method & RAW & ROI & ROI+M & DAE & HOG & LBP  \\
  \midrule
         &  CNN+MLP	& $0.51(\pm 0.04)$ & $0.63(\pm 0.06)$ & $0.65(\pm 0.03)$ & $0.50(\pm 0.07)$ & $0.38(\pm 0.13)$ & $0.25(\pm 0.14)$ \\
Three    &  RF	    & $0.51(\pm 0.02)$ & $0.65(\pm 0.04)$ & $0.67(\pm 0.05)$ & $0.66(\pm 0.04)$ & $0.69(\pm 0.02)$ & $0.60(\pm 0.05)$ \\
Classes  &  KNN	    & $0.48(\pm 0.10)$ & $0.54(\pm 0.02)$ & $0.58(\pm 0.05)$ & $0.55(\pm 0.06)$ & $0.59(\pm 0.02)$ & $0.43(\pm 0.04)$ \\
         &  SVM	    & $0.56(\pm 0.04)$ & $0.66(\pm 0.05)$ & \textbf{0.70}($\pm$ \textbf{0.03}) & $0.70(\pm 0.04)$ & $0.53(\pm 0.02)$ & $0.25(\pm 0.15)$ \\
  \midrule
         &  CNN+MLP	& $0.59(\pm 0.05)$ & $0.80(\pm 0.06)$ & $0.80(\pm 0.03)$ & $0.74(\pm 0.05)$ & $0.72(\pm 0.03)$ & $0.53(\pm 0.03)$ \\
Cascaded &  RF	    & $0.56(\pm 0.05)$ & $0.76(\pm 0.03)$ & $0.74(\pm 0.03)$ & $0.73(\pm 0.03)$ & \textbf{0.81}($\pm$ \textbf{0.03}) & $0.64(\pm 0.09)$ \\
Two      &  KNN	    & $0.47(\pm 0.04)$ & $0.60(\pm 0.06)$ & $0.64(\pm 0.10)$ & $0.56(\pm 0.06)$ & $0.70(\pm 0.03)$ & $0.68(\pm 0.05)$ \\
Classes  &  SVM	    & $0.56(\pm 0.02)$ & $0.78(\pm 0.06)$ & $0.71(\pm 0.05)$ & $0.77(\pm 0.02)$ & $0.74(\pm 0.05)$ & $0.52(\pm 0.00)$ \\
\midrule
         &  CNN+MLP	& $0.55(\pm 0.01)$ & \textbf{0.72}($\pm$ \textbf{0.05}) & $0.70(\pm 0.04)$ & $0.66(\pm 0.05)$ & $0.54(\pm 0.02)$ & $0.21(\pm 0.13)$ \\
Cascaded &  RF	    & $0.47(\pm 0.07)$ & $0.67(\pm 0.03)$ & $0.64(\pm 0.03)$ & $0.65(\pm 0.04)$ & $0.70(\pm 0.03)$ & $0.56(\pm 0.07)$ \\
Final      &  KNN	& $0.44(\pm 0.04)$ & $0.55(\pm 0.06)$ & $0.56(\pm 0.07)$ & $0.52(\pm 0.08)$ & $0.60(\pm 0.05)$ & $0.48(\pm 0.08)$ \\
          &  SVM	& $0.38(\pm 0.02)$ & $0.51(\pm 0.04)$ & $0.46(\pm 0.04)$ & $0.51(\pm 0.04)$ & $0.46(\pm 0.04)$ & $0.10(\pm 0.00)$ \\

  \bottomrule
\end{tabularx}
}
\end{table}

\section{Summary and Conclusion}
In this work, we present a deep learning approach towards grading of collateral flow in ischemic stroke patient based on parametric information extracted from MR perfusion data. We start by extracting region of interest using deep reinforcement learning. We then learn denoising auto-encoder features and modern implementation of 3-D HOG and LBP features. We proceed to the actual classification task using a combination of the extracted features and CNN, random forest, K-nearest neighbour and support vector machine classifiers.

Our experiments show that the rich information on blood flow visible from MRI perfusion can be used to predict collateral flow in a similar manner to DSA images which are invasive in nature. Region of interest detection with reinforcement learning is successful to an acceptable level and can be used as a guide to estimate the region in the brain which requires more attention. We show that for datasets with high class imbalance, a cascaded classification approach performs better than a one time multi-label classification method. It is also evident from our results that a direct CNN classifier is able to extract relevant features from the region of interest and has an advantage over classical classifiers like RF, KNN, and SVM which depends on handcrafted features like HOG and LBP.

Collateral flow grading is an essential clinical procedure in the treatment of ischemic stroke patients. We have presented a framework for automating the process in clinical setup and have achieve promising results given our limited dataset. For the proposed framework to be clinically useful there is the need for further test with possibly more data from multiple stroke centers. The grading can also be customized for specific patient groups for example providing information about age group, gender and other biographical and historical information of patient as an additional feature can help improve the result of the framework.

We make our code used for each experiment in our work public available on github. This will help foster further research in this direction and make it easy for future improvements in the proposed framework.

\bibliographystyle{frontiersinSCNS_ENG_HUMS} % for Science, Engineering and Humanities and Social Sciences articles, for Humanities and Social Sciences articles please include page numbers in the in-text citations
\bibliography{references}

\end{document}